\documentclass[10pt,twocolumn,letterpaper]{article}

\usepackage{cvpr}
\usepackage{times}
\usepackage{epsfig}
\usepackage{graphicx}
\usepackage{amsmath}
\usepackage{amssymb}

\usepackage{sidecap}
\usepackage{wrapfig}
\usepackage[subrefformat=parens,labelformat=parens]{subcaption} 
\usepackage{caption}
\usepackage{adjustbox}
\usepackage{float}
\usepackage{geometry}

\usepackage{multirow}
\usepackage{graphicx}
\usepackage[acronym]{glossaries}
\usepackage{acronym}
\usepackage{collcell}

\usepackage[pagebackref=true,breaklinks=true,letterpaper=true,colorlinks,bookmarks=false]{hyperref}

\usepackage{booktabs}       
\usepackage [english]{babel}
\usepackage [autostyle, english = american]{csquotes}
\MakeOuterQuote{"}
\usepackage[font=itshape]{quoting}
\usepackage{lipsum}
\usepackage{graphics}
\usepackage{xcolor}
\usepackage{color, colortbl}
\usepackage{comment}
\usepackage{booktabs, ragged2e}
\usepackage{multirow}
\usepackage{makecell}
\usepackage{balance}

\usepackage{pifont}
\usepackage{balance}

\graphicspath{{figures/}} 
\usepackage{bibunits}

\defaultbibliography{bias}
\defaultbibliographystyle{ieee_fullname}

\usepackage[acronym]{glossaries}
\usepackage{acronym}

\usepackage{array}
\newcolumntype{x}[1]{>{\centering\arraybackslash\hspace{0pt}}p{#1}}
\newcommand{\ie}{\textit{i}.\textit{e}., }
\newcommand{\eg}{\textit{e}.\textit{g}., }

\newcommand{\xmark}{\ding{56}}%
\newcommand{\checkc}{\ding{51}}%

\newacronym{ml}{ML}{machine learning}
\newacronym{fr}{FR}{facial recognition}
\newacronym{fv}{FV}{facial verification}

\newacronym{cnn}{CNN}{convolutional neural network}
\newacronym{nn}{NN}{neural network}
\newacronym{mtcnn}{MTCNN}{\emph{multi-task \gls{cnn}}}

\newacronym{gan}{GAN}{generative adversarial network}
\newacronym{se}{SE}{\emph{Squeeze-and-Excitation}}
\newacronym{d}{$D$}{discriminator}
\newacronym{g}{$G$}{generator}
\newacronym{dbvae}{DB-VAE}{Debiasing Variational Autoencoder}

\newacronym{lut}{LUT}{Look-Up-Table}
\newacronym{soa}{SOTA}{state-of-the-art}

\newacronym{fiw}{FIW}{Families In the Wild}
\newacronym{lfw}{LFW}{Labeled Faces in the Wild}
\newacronym{bfw}{BFW}{Balanced Faces In the Wild}
\newacronym{rfw}{RFW}{Racial Faces in-the-Wild:}
\newacronym{dp}{DemogPairs}{Demographic Pairs}
\newacronym{itwcc}{ITWCC}{Wild Child Celebrity}

\newacronym{m}{M}{\textit{Male}}
\newacronym{f}{F}{\textit{Female}}
\newacronym{a}{A}{\textit{Asian}}
\newacronym{b}{B}{\textit{Black}}
\newacronym{i}{I}{\textit{Indian}}
\newacronym{w}{W}{\textit{White}}
\newacronym{af}{AF}{\textit{Asian}-\textit{Female}}
\newacronym{am}{AM}{\textit{Asian}-\textit{Male}}
\newacronym{bf}{BF}{\textit{Black}-\textit{Female}}
\newacronym{bm}{BM}{\textit{Black}-\textit{Male}}
\newacronym{if}{IF}{\textit{Indian}-\textit{Female}}
\newacronym{im}{IM}{\textit{Indian}-\textit{Male}}
\newacronym{wf}{WF}{\textit{White}-\textit{Female}}
\newacronym{wm}{WM}{\textit{White}-\textit{Male}}

\newacronym{fd}{FD}{Face Discrimination}
\newacronym{bb}{BB}{bounding box}

\newacronym{sdm}{SDM}{signal detection model}
\newacronym{roc}{ROC}{receiver operating characteristic}
\newacronym{nmse}{NMSE}{Normalized Mean Square Error}
\newacronym{det}{DET}{Detection error trade-off}
\newacronym{tp}{TP}{true-positive}
\newacronym{fp}{FP}{false-positive}

\newacronym{tpir}{TPIR}{true-positive identification rate}
\newacronym{frir}{FRIR}{false-reject identification rate}
\newacronym{fpir}{FRIR}{false-positive identification rate}

\newacronym{fn}{FN}{false-negative}
\newacronym{frr}{FRR}{false-reject rate}
\newacronym{fnr}{FNR}{false-negative rate}

\newacronym{fpr}{FPR}{false-positive rate}

\newacronym{tpr}{TPR}{true-positive rate}

\newacronym{tar}{TAR}{True Acceptance Rate}
\newacronym{far}{FAR}{False Acceptance Rate}
\newacronym{eer}{EER}{Equal Error Rate}

\newacronym{cs}{CS}{Cosine Similarity}

\newacronym{lime}{LIME}{Local Interpretable Model-Agnostic Explanations}
\newacronym{nas}{NAS}{Neural Architecture Search}
\newacronym{gapf}{GAPF}{Generative Adversarial Privacy and Fairness}

\definecolor{Gray}{gray}{0.85}
\definecolor{LightCyan}{rgb}{0.88,1,1}

\newcolumntype{a}{>{\columncolor{Gray}}c}
\newcolumntype{b}{>{\columncolor{white}}c}

\newcommand{\vo}{\vec{o}\@ifnextchar{^}{\,}{}}

\newcommand{\vx}{\vec{x}\@ifnextchar{^}{\,}{}}

\def\colorModel{hsb} 
\usepackage{babel}

\usepackage{hhline}
\newcommand\ColCell[1]{
  \pgfmathparse{#1<50?1:0}  
    \ifnum\pgfmathresult=0\relax\color{white}\fi
  \pgfmathsetmacro\compA{0}      
  \pgfmathsetmacro\compB{#1/100} 
  \pgfmathsetmacro\compC{1}      
  \edef\x{\noexpand\centering\noexpand\cellcolor[\colorModel]{\compA,\compB,\compC}}\x #1
  } 
\newcolumntype{E}{>{\collectcell\ColCell}m{0.4cm}<{\endcollectcell}}  

\cvprfinalcopy 

\ifcvprfinal\fi
\begin{document}

\title{Face Recognition: Too Bias, or Not Too Bias?}

\author{\parbox{16cm}{\centering
    {\large Joseph P Robinson$^1$, Gennady Livitz$^2$, Yann Henon$^2$, Can Qin$^1$,\\ Yun Fu$^1$, and Samson Timoner$^2$}\\
    {\normalsize
    \hspace{-.4in}$^{1}$Northeastern University\hspace{.7in} $^{2}$ISM Connect}}
}

\maketitle

\begin{abstract}
We reveal critical insights into problems of bias in state-of-the-art \gls{fr} systems using a novel \gls{bfw} dataset: data balanced for gender and ethnic groups. We show variations in the optimal scoring threshold for face-pairs across different subgroups. Thus, the conventional approach of learning a  global threshold for all pairs results in performance gaps between subgroups. By learning subgroup-specific thresholds, we reduce performance gaps, and also show a notable boost in overall performance. Furthermore, we do a human evaluation to measure bias in humans, which supports the hypothesis that an analogous bias exists in human perception. For the \gls{bfw} database, source code, and more, visit \href{https://github.com/visionjo/facerec-bias-bfw}{https://github.com/visionjo/facerec-bias-bfw}.
\end{abstract}


\glsresetall
\section{Introduction}
    As more of society becomes integrated with \gls{ml}, topics related to bias, fairness, and the formalization of \gls{ml} standards attract more attention~\cite{10.1007/978-3-030-13469-3_68, anne2018women, wang2018racial}. Thus, an effect of the growing dependency on \gls{ml} is an ever-increasing concern about the biased and unfair algorithms, \eg untrustworthy and prejudiced \gls{fr}~\cite{nagpal2019deep,snow2018}.

    Nowadays, \glspl{cnn} are trained on faces identified by a detection system. Specifically, for \gls{fr}, the goal is to encode faces in an N-dimensional space such that samples of the same identity are neighbors, while different persons are farther apart. Thus, we can deploy a \gls{cnn} to encode faces, and then compare via a similarity score (\ie pairs with a high enough score classified as \emph{genuine}; else, \emph{imposter}). 

\begin{figure}
    \centering
    \includegraphics[width=\linewidth]{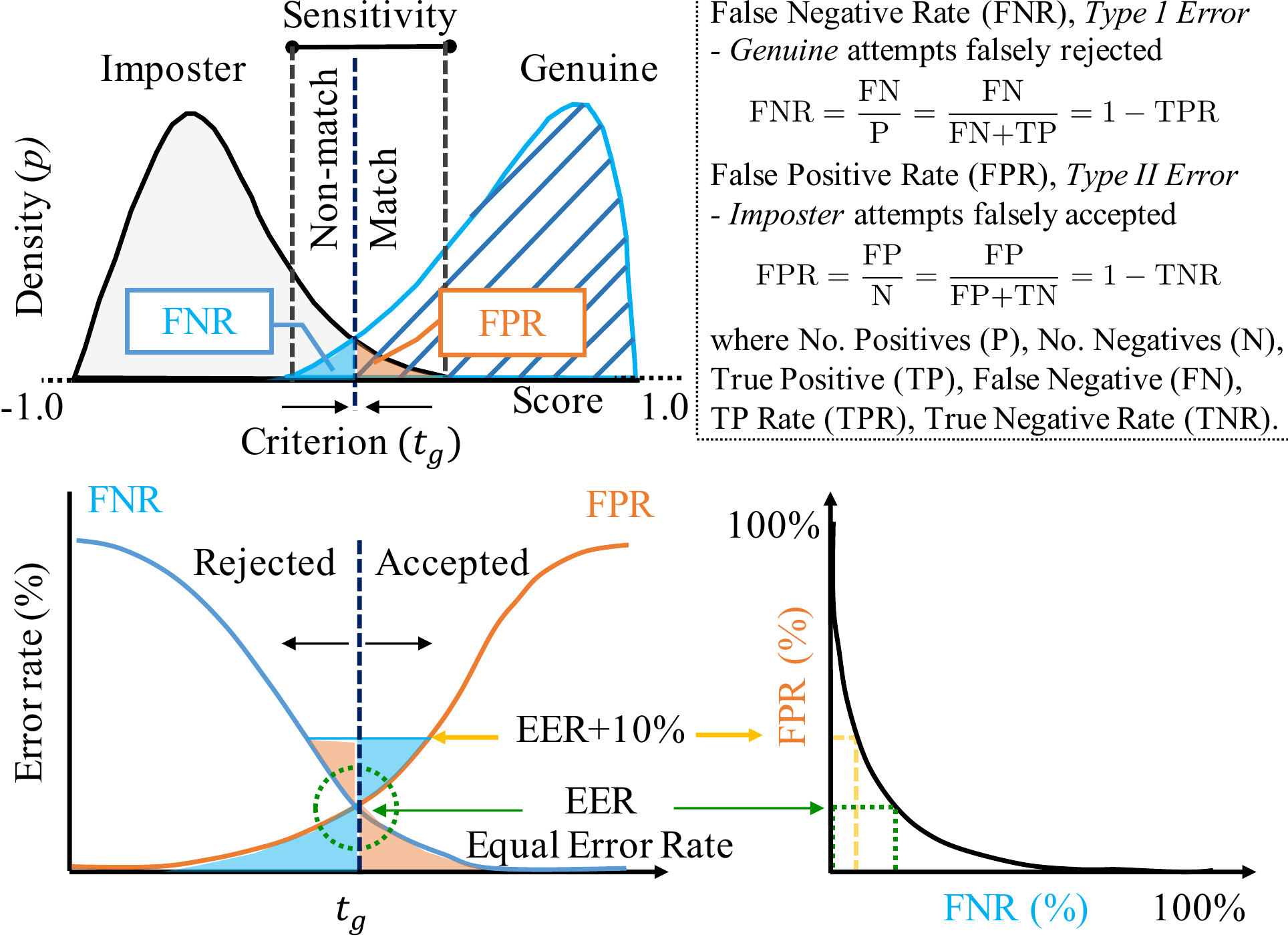}
    \caption{\small{\textbf{Depiction of bio-metrics.} The \gls{sdm} (\emph{top-left}) shows the sensitivity related to a single threshold $t_g$. Translated to error rate (\emph{bottom-left}), a direct trade-off between FNP and FPR. In practice, this is entirely application dependent. Specifically, the specific chose in desired FPR (\%) (\emph{bottom-right}).}}
    \label{fig:biometrics}
\end{figure}

Typically, a fixed threshold sets the decision boundary by which to compare scores (Fig.~\ref{fig:biometrics}). As such, features of the same identity must satisfy a criterion based on a single value~\cite{deng2019arcface, liu2017sphereface, wang2018additive, wang2018cosface}. However, we found that an individual (\ie global) threshold is a crude measure that leads to skewed errors. Furthermore, the held-out set used to determine the threshold tends to share the same distribution with the test set, favoring specific demographics that are the majority. That skew (\ie the difference in the performance of an algorithm of particular demographics) is our definition of bias. A key question is: \emph{is \gls{fr} too biased, or not?}

\begin{table*}[!t]
    \centering
    \caption{\small{\textbf{Database stats and nomenclature.} \textit{Header:} Subgroup definitions. \textit{Top:} Statistics of \gls{bfw}. \textit{Bottom:} Number of pairs for each partition. Columns grouped by ethnicity and then further split by gender.}}\label{tab:ethnic-splits}
    \scalebox{.75}{
     \resizebox{\textwidth}{!}{%
    \begin{tabular}{r c c c c c c c c l}
        \toprule
        & \multicolumn{2}{c}{Asian (A)} & \multicolumn{2}{c}{Black (B)}  & \multicolumn{2}{c}{Indian (I)}& \multicolumn{2}{c}{White (W)}\\
        \cmidrule(l){2-3} \cmidrule(l){4-5} \cmidrule(l){6-7}\cmidrule(l){8-9} 
         & Female (AF) & Male (AM) & BF & BM& IF & IM & WF & WM&Aggregated\\ 
        \midrule

       \# Faces  &  2,500&  2,500& 2,500 & 2,500& 2,500 & 2,500 & 2,500 & 2,500 &20,000 \\ 
        \# Subjects & 100& 100& 100  & 100  & 100  & 100& 100 &100&800  \\ 
        \# Faces / Subject  & 25 & 25    & 25 & 25 & 25  & 25  &  25 & 25 & 25\\ 
\specialrule{.01em}{.05em}{.05em}
            \# Positive Pairs &  30,000&  30,000& 30,000 &30,000 & 30,000 &30,000&30,000 & 30,000 &240,000 \\ 
        \# Negative Pairs & 85,135&  85,232& 85,016  & 85,141  & 85,287  & 85,152& 85,223 &85,193&681,379  \\ 

        \# Pairs (Total) & 115,135 & 115,232    &115016 &115,141 & 115287  & 115,152  &  115,223& 115193 & 921,379\\ 
        
        \bottomrule
    \end{tabular}}
    }
    \vspace{-12pt}
\end{table*}

\begin{figure}[t!]
    \centering
    \includegraphics[width=.8\linewidth]{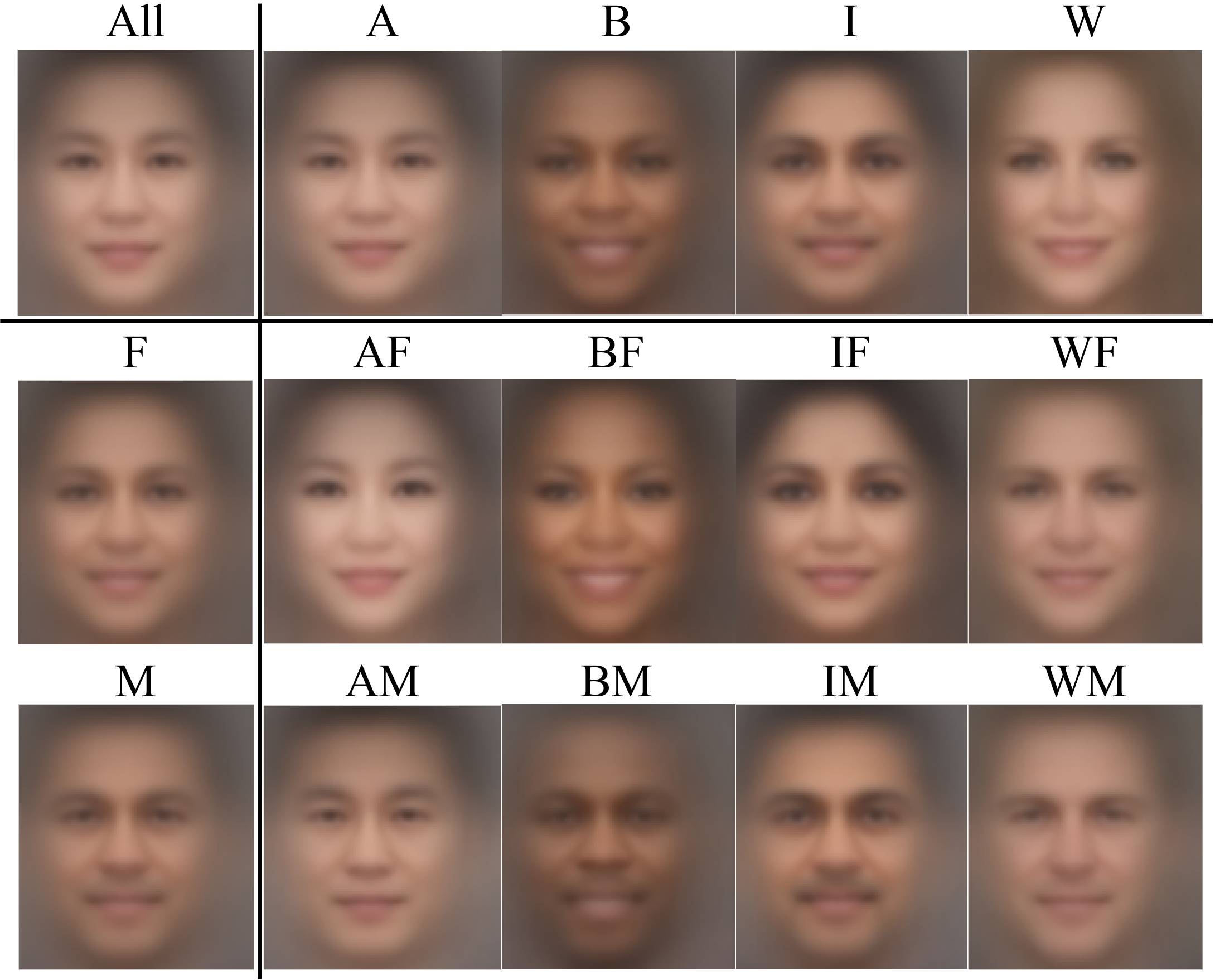}
    \caption{\small{\textbf{\gls{bfw}.} Average face per subgroup: \emph{top-left}: the entire \gls{bfw}; \emph{top-row} per ethnicity;  \emph{left-column}: per gender. The others represent the ethnicity and gender, respectively. Table~\ref{tab:ethnic-splits} defines the acronyms of subgroups.}}
    \label{fig:avg-faces}
\end{figure}

    Making matters more challenging, precise definition of race and ethnicity vary from source-to-source. For example, the US Census Bureau allows an individual to self-identify race.\footnote{\scriptsize\href{https://www.census.gov/mso/www/training/pdf/race-ethnicity-onepager.pdf}{www.census.gov/mso/www/training/pdf/race-ethnicity-onepager}} Even gender, our attempt to encapsulate the complexities of the sex of a human as one of two labels. Others have addressed the oversimplified class labels by representing gender as a continuous value between 0 and 1 - rarely is a person entirely \emph{M} or \emph{F}, but most are somewhere in between~\cite{merler2019diversity}. For this work, we define subgroups as specific sub-populations with face characteristics similar to others in a region. Specifically, we focus on 8 subgroups (Fig.~\ref{fig:avg-faces}).

    The adverse effects of a global threshold are two-fold: \textbf{(1)} mappings produced by \glspl{cnn} are nonuniform. Therefore, distances between pairs of faces in different demographics vary in distribution of similarity scores (Fig~\ref{fig:detection-model}); \textbf{(2)} evaluation set is imbalanced. Subgroups that make up a majority of the population will carry most weight on the reported performance ratings. Reported results favor the common traits over the underrepresented. Demographics like gender, ethnicity, race, and age are underrepresented in most public datasets~\cite{merler2019diversity, wang2018racial}.

    For \textbf{(1)}, we propose subgroup-specific (\ie optimal) thresholds while addressing \textbf{(2)} with a new benchmark dataset to measure bias in \gls{fr}, \gls{bfw} (Table~\ref{tab:ethnic-splits} and~\ref{tab:compared}). \gls{bfw} serves as a proxy for fair evaluations for \gls{fr} while enabling per subgroup ratings to be reported. We use \gls{bfw} to gain an understanding of the extent to which bias is present in \gls{soa} \gls{cnn}s used \gls{fr}. Then, we suggest a mechanism to mitigate problems of bias with more balanced performance ratings for different demographics. Specifically, we propose using an adaptive threshold that varies depending on the characteristics of detected facial attributes (\ie gender and ethnicity). We show an increase in accuracy with a balanced performance for different subgroups. Similarly, we show a positive effect of adjusting the similarity threshold based on the facial features of matched faces. Thus, selective use of similarity thresholds in current \gls{soa} \gls{fr} systems provides more intuition in \gls{fr} research with a method easy to adopt in practice.

    The contributions of this work are 3-fold. (1) We built a balanced dataset as a proxy to measure verification performance per subgroup for studies of bias in \gls{fr}. (2) We revealed an unwanted bias in scores of face pairs - a bias that causes ratings to skew across demographics. For this, we showed that an adaptive threshold per subgroup balances performance (\ie the typical use of a global threshold unfavorable, which we address via optimal thresholds). (3) We surveyed humans to demonstrate bias in human perception (NIH-certified, \textit{Protect Humans in Research}).

\section{Background Information}
    \subsection{Bias in \gls{ml}}
        The progress and commercial value of \gls{ml} are exciting. However, due to inherent biases in ML, society is not readily able to trust completely in its widespread use. The exact definitions and implications of bias vary between sources, as do its causes and types. A common theme is that bias hinders performance ratings in ways that skew to a particular sub-population. In essence, the source varies, whether from humans~\cite{windmann1998subconscious}, data or label types~\cite{tommasi2017deeper}, \gls{ml} models~\cite{amini2019uncovering, kim2019learning}, or evaluation protocols~\cite{stock2018convnets}. For instance, a vehicle-detection model might miss cars if training data were mostly trucks. In practice, many \gls{ml} systems learn biased data, which could be detrimental to society.

    \subsection{Bias in \gls{fr}}
        Different motivations have driven problems of bias in \gls{fr}. Instances can be in issues of data augmentation~\cite{yin2019feature}, one-shot learning~\cite{ding2018one}, demographic parity and fairness with priority on privacy~\cite{huang2018generative}, domain adaptation~\cite{wang2018racial}, differences in face-based attributes across demographics~\cite{wang2018they}, data exploration~\cite{muthukumar2019}, and even characterizing different commercial systems~\cite{buolamwini2018gender}. 
        
        Yin~\etal proposed to augment the feature space of underrepresented classes using different classes with a diverse collection of samples~\cite{yin2019feature}. This was to encourage distributions of underrepresented classes to resemble the others more closely. Similarly, others formulated the imbalanced class problem as one-shot learning, where a \gls{gan} was trained to generate face features to augment classes with fewer samples~\cite{ding2018one}. \gls{gapf} was proposed to create fair representations of the data in a quantifiable way, allowing for the finding of a de-correlation scheme from the data without access to its statistics~\cite{huang2018generative}. Wang~\etal defined subgroups at a finer level (\ie Chinese, Japanese, Korean), and determined the familiarity of faces inter-subgroup~\cite{wang2018they}. Genders have also been used to make subgroups (\eg for analysis of gender-based face encodings~\cite{muthukumar2019}). Most recently,~\cite{wang2018racial} proposed to adapt domains to bridge the bias gap by knowledge transfer, which was supported by a novel data collection, \gls{rfw}. The release of \gls{rfw} occurred after \gls{bfw} was built - although similar in terms of demographics, \gls{rfw} uses faces from MSCeleb~\cite{guo2016ms} for testing, and assumes CASIA-Face~\cite{yi2014learning} and VGG2~\cite{Cao18} were used to train. In contrast, our \gls{bfw} assumes VGG2 as the test set. 
        Furthermore, \gls{bfw} balance subgroups: \gls{rfw} splits subgroups by gender and race, while \gls{bfw} has gender, race, or both). 

Most similar to us is~\cite{das2018, demogPairs, lopez2019dataset, srinivas2019face} - each was motivated by insufficient paired data for studying bias in \gls{fr}. Then, problems were addressed using labeled data from existing image collections. Uniquely, Hupont~\etal curated a set of faces based on racial demographics (\ie \gls{a}, \gls{b}, and \gls{w}) called \gls{dp}~\cite{demogPairs}. In contrast,~\cite{srinivas2019face} honed in on adults versus children called \gls{itwcc}. Like the proposed \gls{bfw}, both were built by sampling existing databases, but with the addition of tags for the respective subgroups of interest. Aside from the additional data of \gls{bfw} (\ie added subgroup \gls{i}, along with other subjects with more faces for all subgroups), we also further split subgroups by gender. Furthermore, we focus on the problem of facial verification and the different levels of sensitivity in cosine similarity scores per subgroup.

    \begin{table*}[t!]
        
        \centering
        \caption{\small{\textbf{\gls{bfw} and related datasets.} \gls{bfw} is balanced across ID, gender, and ethnicity (Table~\ref{tab:ethnic-splits}). Compared with \gls{dp}, \gls{bfw} provides more samples per subject and subgroups per set, while using a single resource, VGG2. \gls{rfw}, on the other hand, supports domain adaptation, and focuses on race-distribution - not the distribution of identities.}}
        \scriptsize
        \begin{tabular}{rccccccc}
        
            \multicolumn{2}{c}{Database} & \multicolumn{3}{c}{Number of}& \multicolumn{3}{c}{Balanced Labels}\\
            \cmidrule(lr){1-2}	\cmidrule(lr){3-5} \cmidrule(lr){6-8}
            Name & Source Data & Faces &  IDs & Subgroups & ID & Ethnicity & Gender\\\midrule
            \gls{dp}~\cite{demogPairs}     & CASIA-W~\cite{yi2014learning}, VGG~\cite{schroff2015facenet} \&VGG2~\cite{Cao18} & 10,800& 600 & 6 &\checkc& \checkc &\checkc \\
            \gls{rfw}~\cite{wang2018racial}     &  MS-Celeb-1M &$\approx$80,000&$\approx$12,000& 4 & \xmark & \checkc &\xmark \\
            \gls{bfw} (ours) & VGG2 & 20,000 & 800 &8 & \checkc & \checkc &\checkc \\\bottomrule
        \end{tabular}
        \label{tab:compared}
            \vspace{-4mm}
    \end{table*}
    
\subsection{Human bias in \gls{ml}}
Bias is not unique to \gls{ml} - humans are also susceptible to a perceived bias. Biases exist across race, gender, and even age~\cite{10.1007/978-3-030-13469-3_68, bar2006, meissner2001, nicholls2018}. Wang~\etal showed machines surpass human performance in discriminating between Japanese, Chinese, or Korean faces by nearly 150\%~\cite{wang2018they}, as humans just pass random (\ie 38.89\% accuracy).

We expect the human bias to skew to their genders and races. For this, we measure human perception with face pairs of different subgroups (Section~\ref{subsec:human-assessment}). The results concur with~\cite{wang2018they}, as we also recorded overall averages below random ($<$50\%). 

\glsunset{if}\glsunset{im}\glsunset{af}\glsunset{am}\glsunset{bf}\glsunset{bm}\glsunset{wf}\glsunset{wm}
\section{The BFW Benchmark and Dataset}
We now discuss the acquisition of \gls{bfw}, protocols, and settings used to survey bias in humans.

\subsection{The data}
Problems of bias in \gls{fr} motivated us to build \gls{bfw}. Inspired by \gls{dp}~\cite{demogPairs}, the data is made up of evenly split subgroups, but with an increase in subgroups (\ie \gls{if} and \gls{im}), subjects per subgroup, and face pairs (Table~\ref{tab:ethnic-splits}).

\vspace{.5mm}
\noindent\textbf{Compiling subject list.} 
Subjects were sampled from VGG2~\cite{Cao18} - unlike others built from multiple sources, \gls{bfw} has fewer potential conflicts in train and test overlap with existing models. To find candidates for the different subgroups, we first parsed the list of names using a pre-trained ethnicity model~\cite{ambekar2009name}. This was then further refined by processing faces using ethnicity~\cite{fu2014learning} and gender~\cite{levi2015age} classifiers. This resulted in hundreds of candidates per subgroup, which allowed us to manually filter 100 subjects per the 8 subgroups.

\vspace{.5mm}
\noindent\textbf{Detecting faces.} Faces were detected using MTCNN~\cite{zhang2016joint}.\footnote{\href{https://github.com/polarisZhao/mtcnn-pytorch}{https://github.com/polarisZhao/mtcnn-pytorch}} Then, faces were assigned to one of two sets. Faces within detected bounding box (BB) regions extended out 130\% in each direction, with zero-padding as the boundary condition made-up one set. The second set were faces aligned and cropped for Sphereface~\cite{liu2017sphereface} (see the next step). Also, coordinates of the BB and the five landmarks from \gls{mtcnn} were stored as part of the static, raw data. For samples with multiple face detections, we used the BB area times the confidence score of the \gls{mtcnn} to determine the face most likely to be the subject of interest, with the others set aside and labeled \textit{miss-detection}.

\vspace{.5mm}
\noindent\textbf{Validating labels.} 
Faces of \gls{bfw} were encoded using the original implementation of the \gls{soa} Sphereface~\cite{liu2017sphereface}. For this, each face was aligned to predefined eye locations via an affine transformation. Then, faces were fed through the \gls{cnn} twice (\ie the original and horizontally flipped), with two features fused by average pooling (\ie 512 D). A matrix of cosine similarity scores was then generated for each subject and removed samples (\ie rows) with median scores below threshold $\theta=0.2$ (set manually). Mathematically, the $n^{th}$ sample for the $j^{th}$ subject with $N_j$ faces was removed if the ordinal rank of its score $n = \frac{P\times N}{100}\geq\theta$, where $P=50$. In other words, the median (\ie 50 percentile) of all scores for a face with respect to all of faces for the respective subject must pass a threshold of $\theta=0.2$; else, the face is dropped. This allowed us to quickly prune \gls{fp} face detections. Following~\cite{robinson2016families, robinson2018visual}, we built a JAVA tool to visually validate the remaining faces. For this, the faces were ordered by decreasing confidence, with confidence set as the average score, and then displayed as image icons on top toggling buttons arranged as a grid in a sliding pane window. Labeling then consisted of going subject-by-subject and flagging faces of \emph{imposters}.

\begin{figure}[!t] 
	\centering    
 \glsunset{fpr}
  \glsunset{fnr}
	\includegraphics[width=.83\linewidth]{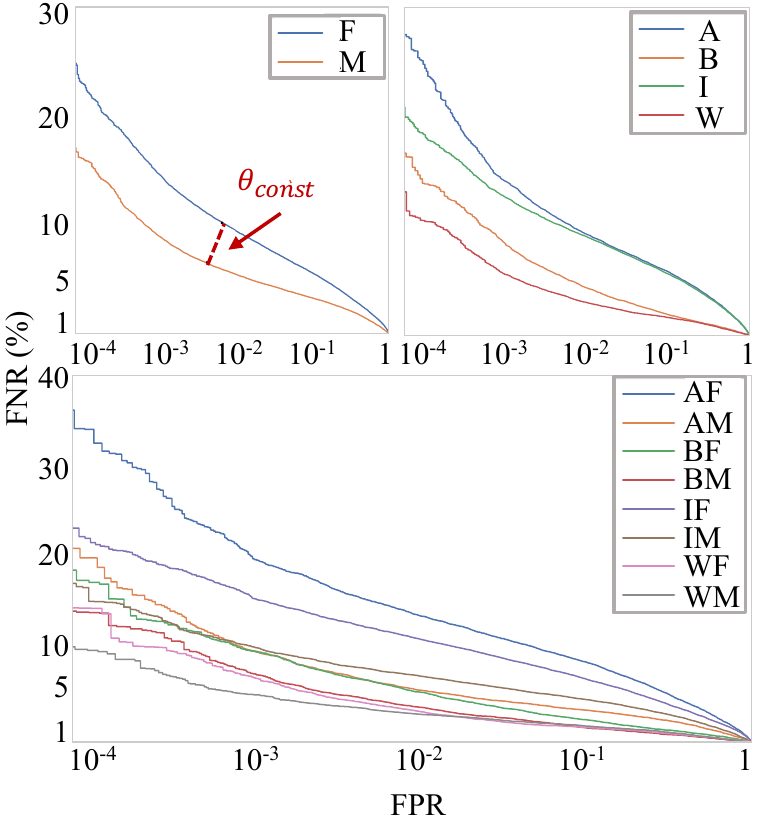}
		\caption{\small{\textbf{\gls{det} curves.} \emph{Top-left}: per gender. \emph{Top-right}: per ethnicity. \emph{Bottom}: per subgroup (\ie combined). Dashed line shows about 2$\times$ difference in \gls{fpr} for the same threshold $\theta_{const}$. \gls{fnr} is the match error count (closer to the bottom is better).}}
		\glsreset{det}\glsreset{fpr}
\label{fig:detcurves} 
 \vspace{-4mm}
\end{figure} 
\noindent\textbf{Sampling faces and creating folds.} We created lists of pairs in five-folds with subjects split evenly per person and without overlap across folds. Furthermore, a balance in the number of faces per subgroup was obtained by sampling twenty-five faces at random from each. Next, we generated a list of all the face pairs per subject, resulting in $\sum_{l=1}^{L}\sum_{k=1}^{K_d} {N_k \choose 2}$ positive pairs, where the number of faces of all $K_l$ subjects $N_k=25$  for each of the $L$ subgroups (Table~\ref{tab:ethnic-splits}). Next, we assigned subjects to a fold. To preserve balance across folds, we sorted subjects by the number of pairs and then started assigning to alternating folds from the one with the most samples. Note, this left no overlap in identity between folds. Later, a negative set from samples within the same subgroup randomly matched until the count met that of the positive. Finally, we doubled the number with negative pairs from across subgroups but in the same fold.

\begin{figure}[t!] 
	\glsunset{fpr}
	\centering
	\centering
	\includegraphics[width=1\linewidth]{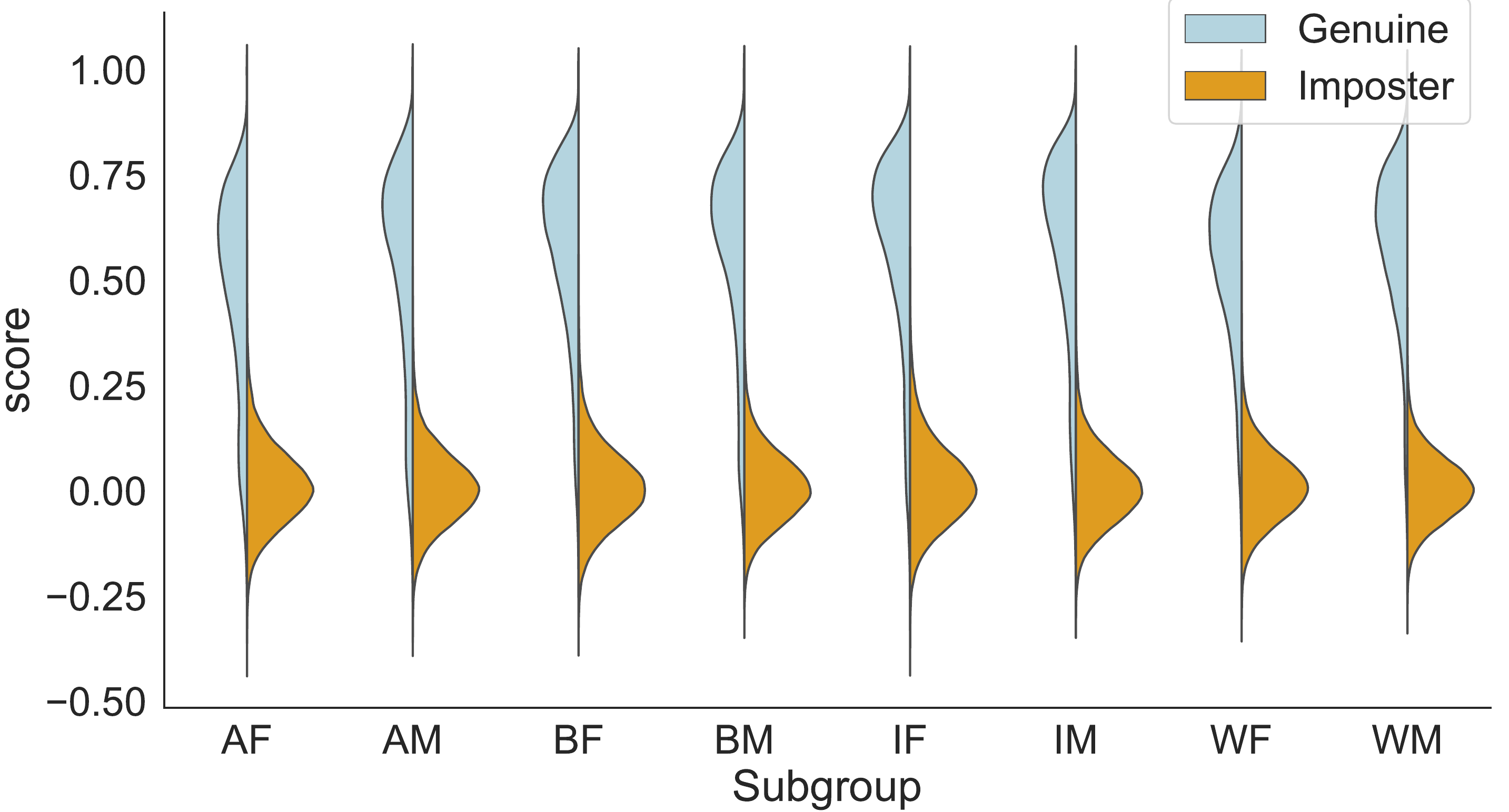}
		\caption{\small{\textbf{\Gls{sdm} across subgroups.} Scores of \emph{imposters} have medians around 0.3 but with variations in upper percentiles; \emph{genuine} pairs vary in mean and spread (\eg \gls{af} has more area of overlap). A threshold varying across different subgroups yields a constant \gls{fpr}.}} \label{fig:detection-model} 
	    \vspace{-4mm}
\end{figure} 

\subsection{Problem formulation}\label{subsec:pf} 
\Gls{fv} is the special case of the two-class (\ie boolean) classification. Hence, pairs are labeled as the ``same'' or ``different'' \textit{genuine} pairs (\ie \textit{match}) or \textit{imposter} (\ie \textit{mismatch}), respectively. This formulation (\ie \gls{fv}) is highly practical for applications like access control, re-identification, and surveillance. Typically, training a separate model for each unique subject is unfeasible. Firstly, the computational costs compound as the number of subjects increase.  Secondly, such a scheme would require model retraining each time a new person is added. Instead, we train models to encode facial images in a feature space that captures the uniqueness of a face, to then determine the outcome based on the output of a scoring (or distance) function. Formally put:
\begin{equation}\label{eg:matcher}
    f_{boolean}(\vec{x}_i, \vec{x}_j) = d(\vec{x}_i, \vec{x}_j) \leq \theta,
\end{equation}

where $f_{boolean}$ is the \textit{matcher} of the feature vector $\vec{x}$ for the $i^{th}$ and $j^{th}$ sample~\cite{LFWTech}.

Cosine similarity is used as the \emph{matcher} in Eq~\ref{eg:matcher} the closeness of $i^{th}$ and $j^{th}$ features, \ie
$
s_l= \frac{f_i\cdot f_j}{||f_i||_2||f_j||_2}
$ is the closeness of the $l^{th}$ pair. 

\begin{figure}[t!]
	\centering    
	\includegraphics[width=.65\linewidth]{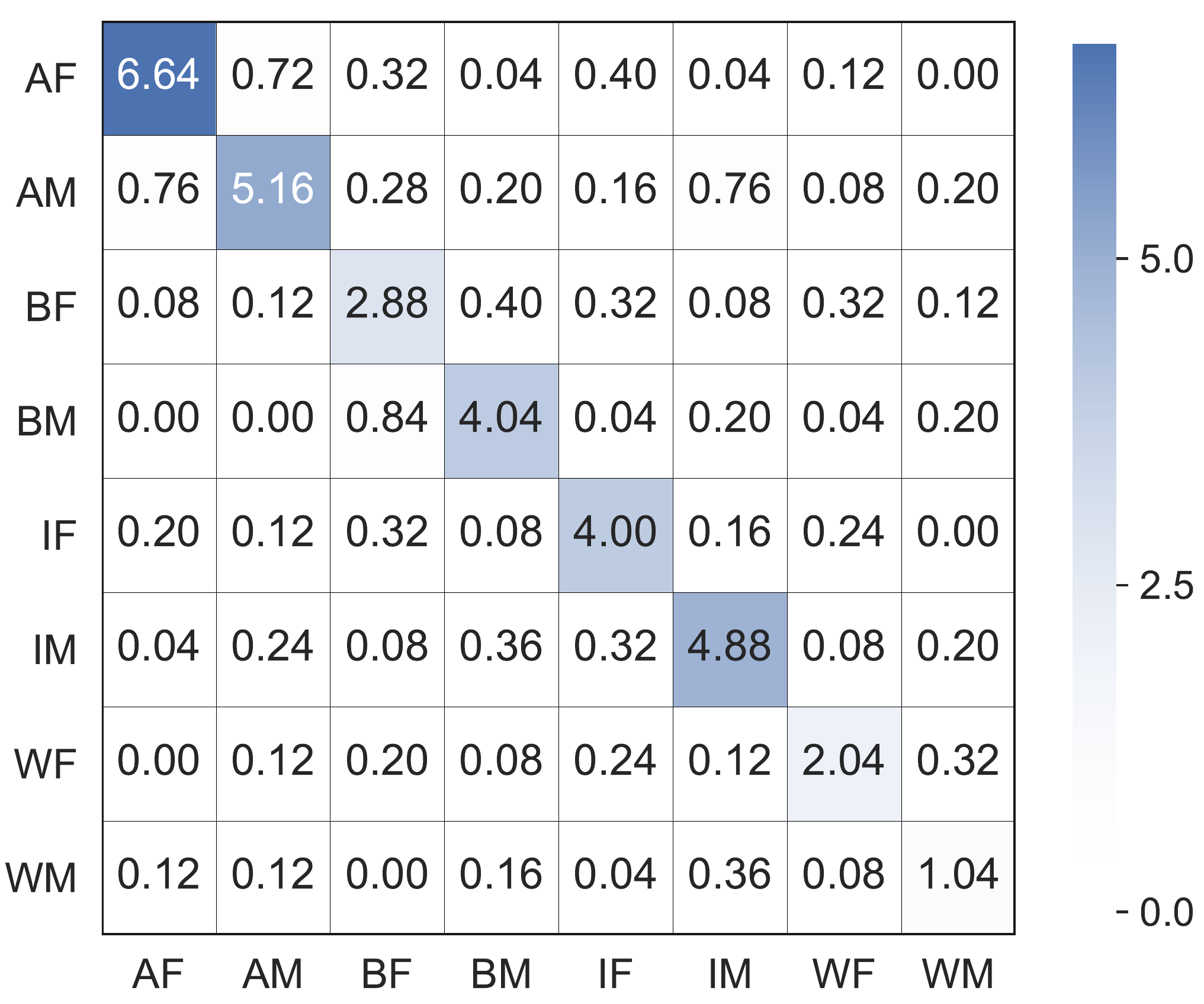}
		\caption{\small{\textbf{Confusion matrix.} Error (Rank 1, \%) for all \gls{bfw} faces versus all others. Errors concentrate intra-subgroup - consistent with the \gls{sdm} (Fig.~\ref{fig:detection-model}). Although subgroups are challenging to define, this shows the ones chosen are meaningful for \gls{fr}.}}
		\label{fig:confusion} 
\end{figure} 

\subsection{Human assessment}\label{subsec:human-assessment}
To focus on the human evaluation experiment, we honed-in on pairs from two groups, White Americans (W) and Chinese from China (C). This minimized variability compared to the broader groups of Whites and Asians, which was thought to be best, provided only a small subset of the data was used on fewer humans than subjects in \gls{bfw}.


Samples were collected from individuals recruited from multiple sources (\eg social media, email lists, family, and friends) - a total of 120 participants were sampled at random from all submissions that were completed and done by a W or C participant. Specifically, there were 60 W and 60 C, both with \gls{m} and \gls{f} split evenly. A total of 50 face pairs of non-famous ``look-alikes'' were collected from the internet, with 20 ({\emph WA}) and 20 ({\emph C}) pairs with, again, \gls{m} and \gls{f} split evenly. The other 10 were of a different demographic (\eg Hispanic/ Latino, Japanese, African). The survey was created, distributed, and recorded via \emph{\href{https://paperform.co}{PaperForm}}. It is important to note that participants were only aware of the task (\ie to verify whether or not a face-pair was a \emph{match} or \emph{non-match}, but with no knowledge of it being a study on the bias).

\begin{figure}[t!]
       \centering
    \includegraphics[width=.95\linewidth]{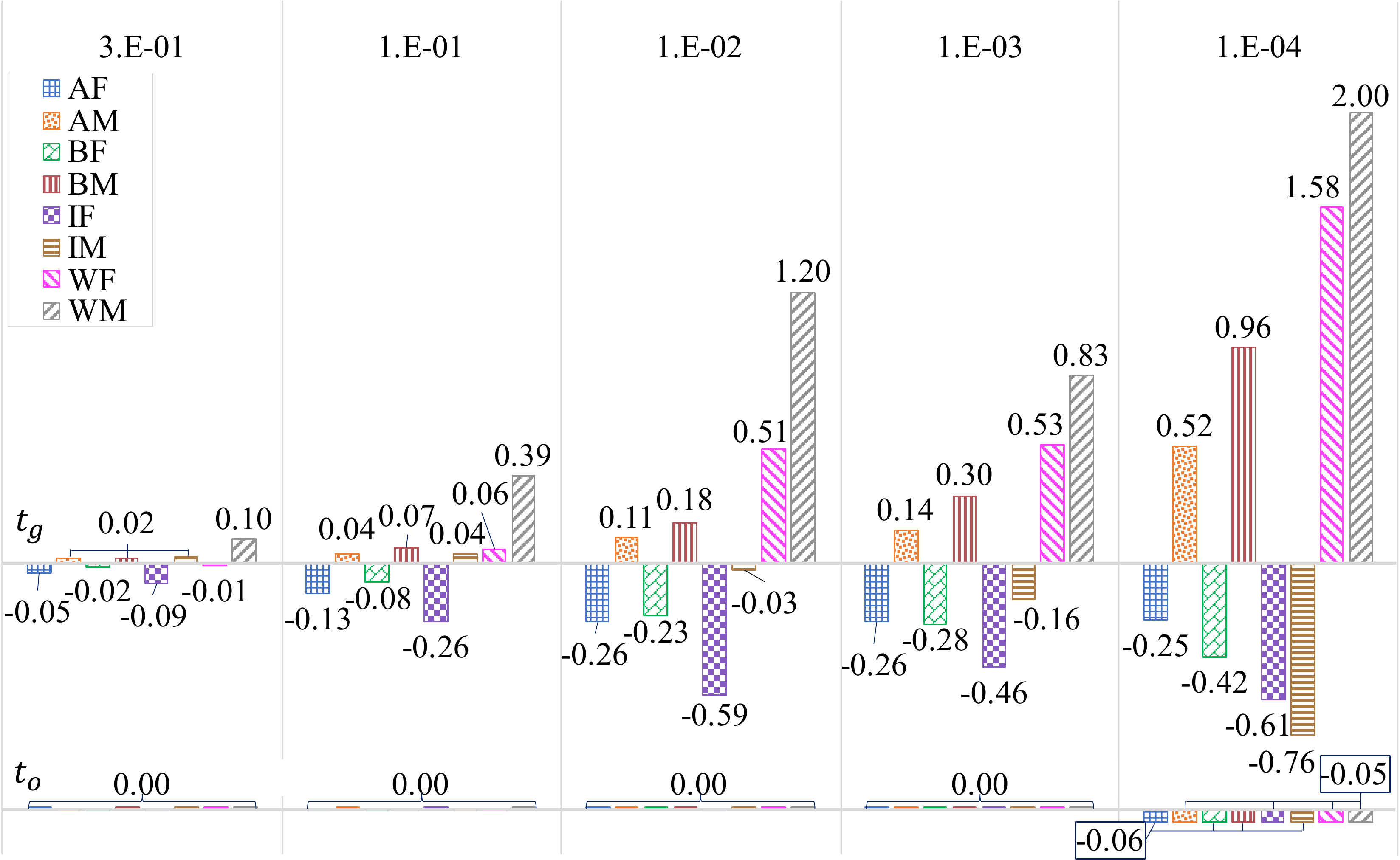}
    \caption{\small{\textbf{Percent difference from intended \gls{fpr}.} \emph{Top:} $t_g$ yields \gls{fpr} the span as large as 2$\times$ (\ie 200\%) that intended (\ie \gls{wm} for 1e-4). Furthermore, \gls{f} subgroups tend to perform worse than intended for all cases (while \gls{m} tend to overshoot intended performance, with exception of \gls{im} in for \gls{fpr}=1e-4). \emph{Bottom:} Subgroup-specific thresholds reduces this difference to near zero, where there are small differences, the percent difference across different subgroups is fair (\ie FPR=1e-4).}}\label{fig:percent:difference}
    \vspace{-4mm}
\end{figure}

\section{Results and Analysis}
A single \gls{cnn} was used as a means to control the experiments. For this, Sphereface~\cite{liu2017sphereface} trained on CASIA-Web~\cite{yi2014learning}, and evaluated on \gls{lfw}~\cite{LFWTech} (\%99.22 accuracy), encoded all of the faces.\footnote{\href{$https://github.com/clcarwin/sphereface\_pytorch$}{https://github.com/clcarwin/sphereface\_pytorch}} As reported in~\cite{wang2019racial}, \gls{lfw} has about 13\%, 14\%, 3\%, and 70\% ratio in Asian, Black, Indian, and White, respectively. Furthermore, CASIA-Web is even more unbalanced (again, as reported in~\cite{wang2018racial}), with about  3\%, 11\%, 2\%, and 85\% for the same subgroups.
\glsunset{det}
\vspace{-5mm}
\noindent\paragraph{\gls{det} analysis.}
\gls{det} curves (5-fold, averaged) show per-subgroup trade-offs (Fig.~\ref{fig:detcurves}). Note that \gls{m} performs better than \gls{f}, precisely as one would expect from the tails of score-distributions for \emph{genuine} pairs (Fig.~\ref{fig:detection-model}). \Gls{af} and \gls{if} perform the worst.
\glsunset{m}
\glsunset{f}


\glsunset{am}\glsunset{af}\glsunset{bm}\glsunset{bf}\glsunset{im}\glsunset{if}\glsunset{wm}\glsunset{wf}
\begin{figure*}[h!]
    \centering
    \begin{subfigure}[t]{.3\linewidth}
    \includegraphics[width=.71\linewidth]{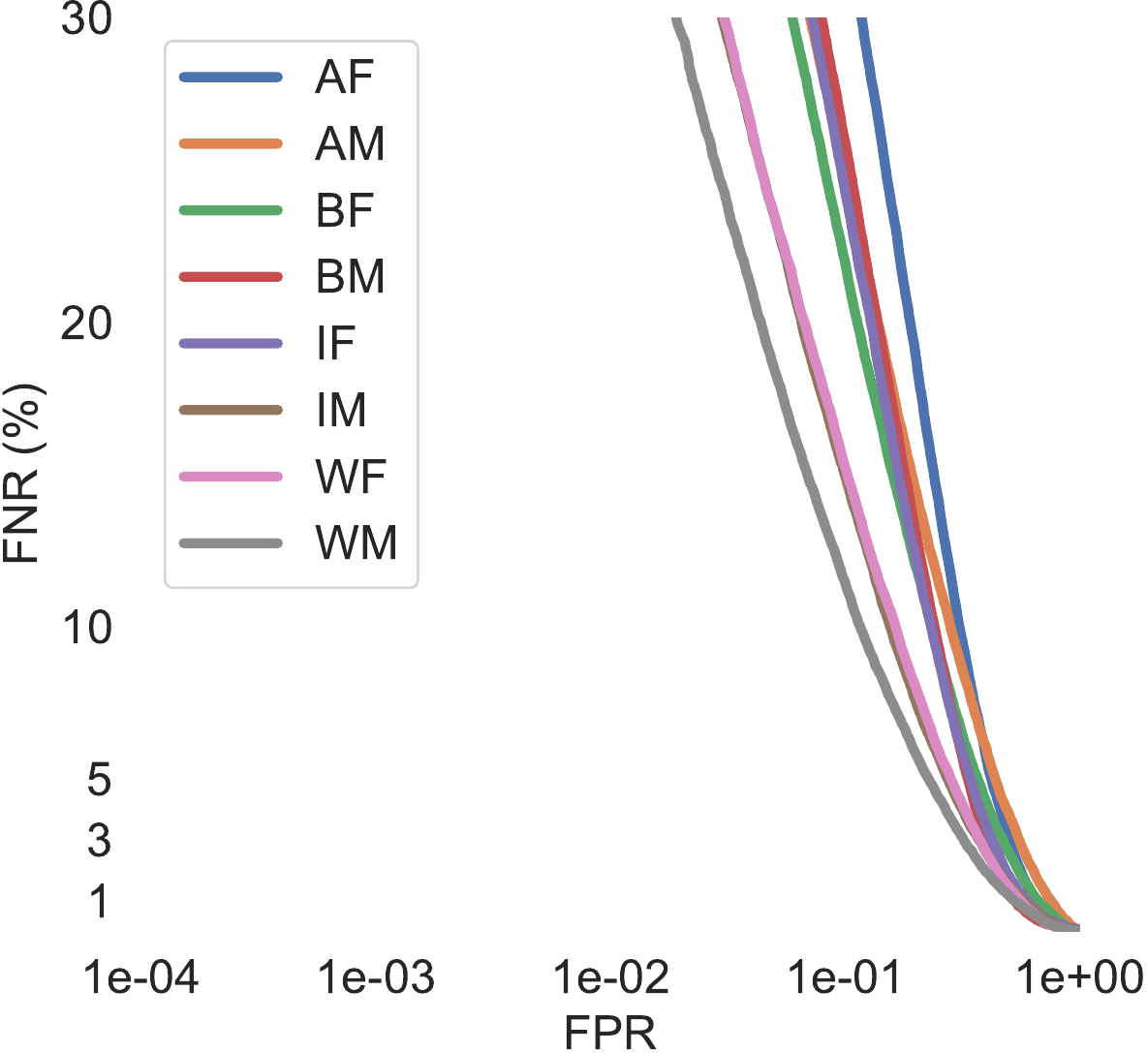}
    \caption{VGG16~\cite{simonyan2014very}}
 \end{subfigure}
    \begin{subfigure}[t]{.27\linewidth}
    \includegraphics[width=.75\linewidth]{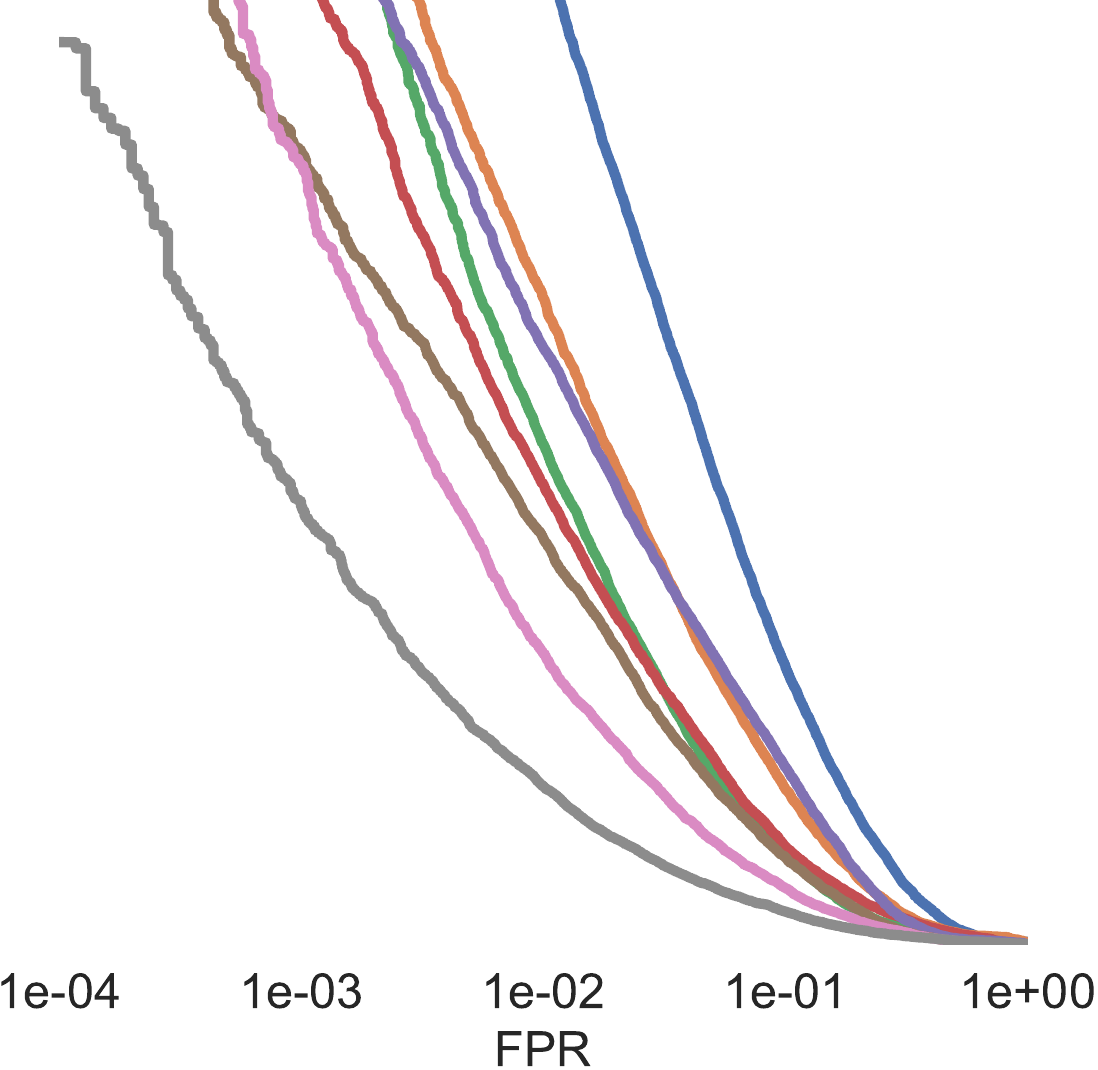}
    \caption{ResNet50~\cite{he2016deep}}
   \end{subfigure}
    \begin{subfigure}[t]{.27\linewidth}
    \includegraphics[width=.75\linewidth]{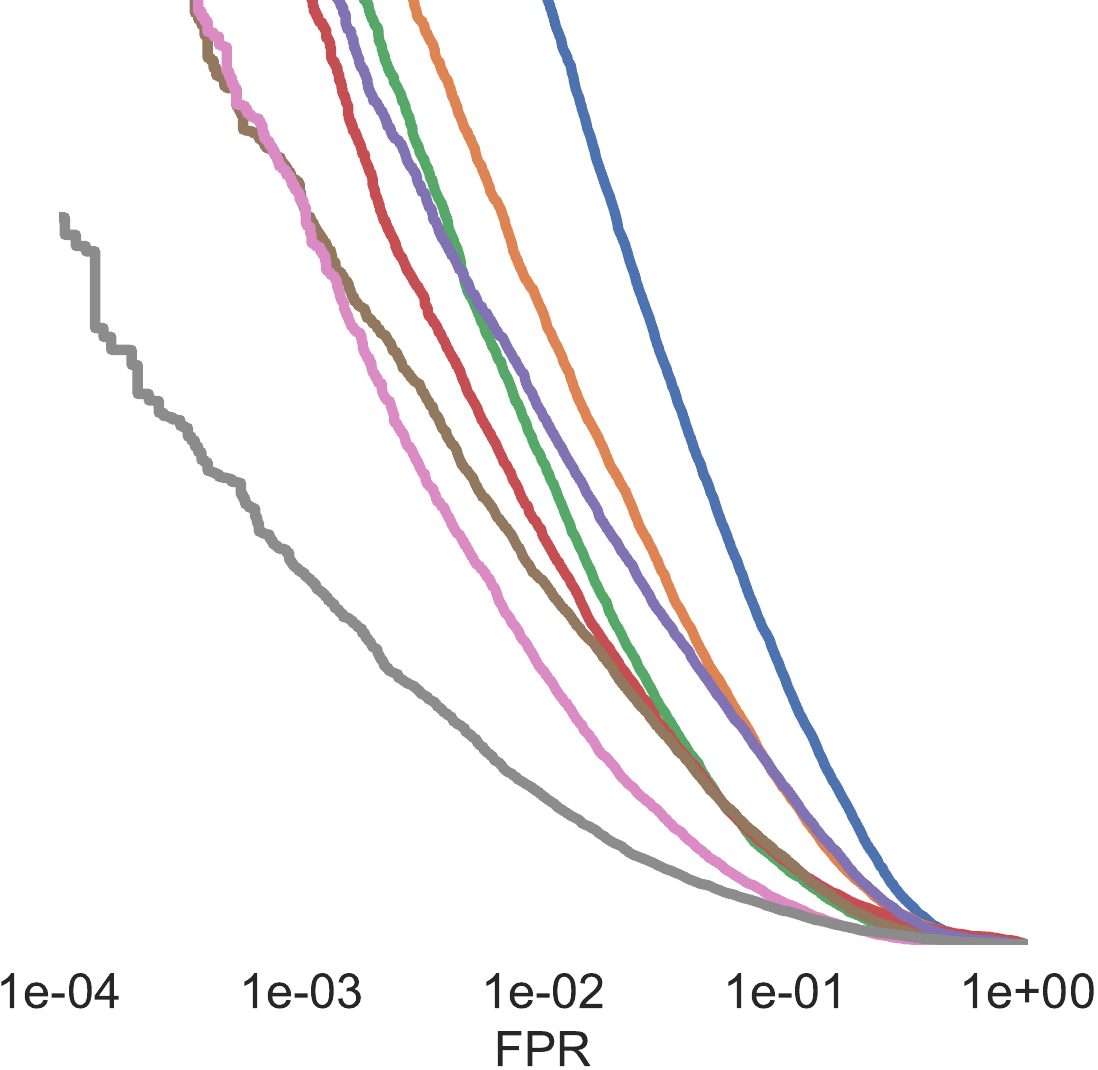}
    \caption{SENet~\cite{hu2018squeeze}}
    \end{subfigure}
    \caption{\small{\textbf{\gls{det} curves for different CNNs}. \gls{fnr} (\%) (vertical) vs \gls{fpr}  (horizontal, log-scale) for VGG2~\cite{Cao18} models with different backbones (VGG16, Resnet50, SEnet50). Lower is better. For each plot, \gls{wm} is the top-performer, while \gls{af} is the worst. The ordering of the curves is roughly the same for each backbone.}}\label{fig:sdm-appendix-a}
    \vspace{-5mm}
\end{figure*}

\vspace{-5mm}
\noindent\paragraph{Score analysis.}
Fig.~\ref{fig:detection-model} shows score distributions for faces of the same (\ie \emph{genuine}) and different (\ie \emph{imposter}) identity, with a subgroup per \gls{sdm} graph. Notice that score distributions for imposters tend to peak about zero for all subgroups, and with minimal deviation comparing modes of the different plots. On the other hand, the score distribution of the \emph{genuine} pairs varies across subgroups in location (\ie score value) and spread (\ie overall shape).
Fig.~\ref{fig:confusion} shows the confusion matrix of the subgroups. A vast majority of errors occur in the intra-subgroup. It is interesting to note that while the definition of  each group  based on ethnicity and race may not be crisply defined, the confusion matrix indicates that in practice, the \gls{cnn} finds that the groups are effectively separate. The categories are, therefore, meaningful for \gls{fr}.


\begin{table}[b!]
    \vspace{-4mm}
\glsunset{tar}
\glsunset{far}
\caption{\small{\textbf{\gls{tar} at intervals of \gls{far}}. \gls{far}, listed are the \gls{tar} scores for a global threshold (top) and the proposed category-based threshold (bottom). Higher is better.}}\label{tab:ethnicy-far} 
\vspace{-2mm}
\begin{center}
\scriptsize
\begin{tabular}{l c c c c c}
     \gls{far} & 0.3 & 0.1 & 0.01 & 0.001 & 0.0001\\\midrule
    \multirow{2}{.1mm}{\textbf{\gls{af}}} &0.990 & 0.867 & 0.516 & 0.470 & 0.465\\[-4pt]
        &1.000 & 0.882 & 0.524 & 0.478 & 0.474\\[-1pt]
    \multirow{2}{3mm}{\textbf{\gls{am}}} &0.994 & 0.883 & 0.529 & 0.482 & 0.477\\[-4pt]
        &1.000 & 0.890 & 0.533 & 0.486 & 0.482\\[-1pt]
    \multirow{2}{3mm}{\textbf{\gls{bf}}} &0.991 & 0.870 & 0.524 & 0.479 & 0.473\\[-4pt]
        &1.000 & 0.879 & 0.530 & 0.484 & 0.480\\[-1pt]
    \multirow{2}{3mm}{\textbf{\gls{bm}}} &0.992 & 0.881 & 0.526 & 0.480 & 0.474\\[-4pt]
        &1.000 & 0.891 & 0.532 & 0.485 & 0.480\\[-1pt]
    \multirow{2}{3mm}{\textbf{\gls{if}}} &0.996 & 0.881 & 0.532 & 0.486 & 0.481\\[-4pt]
        &1.000 & 0.884 & 0.534 & 0.488 & 0.484\\[-1pt]
    \multirow{2}{3mm}{\textbf{\gls{im}}} &0.997 & 0.895 & 0.533 & 0.485 & 0.479\\[-4pt]
        &1.000 & 0.898 & 0.535 & 0.486 & 0.481\\[-1pt]
    \multirow{2}{3mm}{\textbf{\gls{wf}}} &0.988 & 0.878 & 0.517 & 0.469 & 0.464\\[-4pt]
        &1.000 & 0.894 & 0.526 & 0.478 & 0.474\\[-1pt]
    \multirow{2}{3mm}{\textbf{\gls{wm}}} &0.989 & 0.896 & 0.527 & 0.476 & 0.470\\[-4pt]
        &1.000 & 0.910 & 0.535 & 0.483 & 0.478\\[-1pt]
    \midrule
    \multirow{2}{3mm}{\textbf{Avg.}} &0.992 & 0.881 & 0.526 & 0.478 & 0.474\\[-4pt]
        &1.000 & 0.891 & 0.531 & 0.483 & 0.479\\[-10pt]
\end{tabular}
\end{center}
\glsreset{tar}
\glsreset{far}
 \vspace{-3mm}

\end{table}
\glsunset{fpr}

The gender-based \gls{det} curves show performances with a fixed threshold (dashed line). Other curves relate similarity (lines omitted to declutter). For many \gls{fr} applications, systems operate at the highest \gls{fpr} allowed. The constant threshold shows that a single threshold produces different operating points (\ie \gls{fpr}) across subgroups, which is undesirable. 
If this is the case in an industrial system, one would expect a difference in about double the FPs reported based on subgroup alone. The potential ramifications of such a bias should be considered, which it has not as of lately-- even noticed in main-stream media ~\cite{england2019,snow2018}.

To further demonstrate the extent of the problem, we follow settings typical for systems in practice. We set the desired \gls{fpr}, and then determine the percent difference, \ie desired versus actual (Fig.~\ref{fig:percent:difference}, \emph{top}). Furthermore, we mitigate this highly skewed phenomenon by applying subgroup-specific thresholds (\emph{bottom}) - by this, minimal error from the desired \gls{fpr}. Besides where there is a small error, the offset is balanced across subgroups.

\vspace{-5mm}
\noindent\paragraph{Model analysis.}
Variations in optimal threshold exist across models (Fig.~\ref{fig:sdm-appendix-a}). Like in Fig.~\ref{fig:detcurves}, the \gls{det} curves for three \gls{cnn}-based models, each trained on VGG2 with softmax but with different backbones.\footnote{\href{https://github.com/rcmalli/keras-vggface}{https://github.com/rcmalli/keras-vggface}} Notice similar trends across subgroups and models, which is consistent with  Sphereface as well (Fig.~\ref{fig:detcurves}). For example, the plots generated with Sphereface and VggFace2 all have the \gls{wm} curve at the bottom (\ie best) and \gls{af} on top (\ie worst). Thus, the additional \gls{cnn}-based models demonstrate the same phenomena: proportional to the overall model performance, exact in the ordering of curves for each subgroup.

\vspace{-5mm}
\noindent\paragraph{Verification threshold analysis.}
We seek to reduce the bias between subgroups. Such that an operating point (\ie \gls{fpr}) is constant across subgroups. To accomplish that, we used a per subgroup threshold. In \gls{fv}, we consider one image as the query, and all others as the test. For this, the ethnicity of the query image is assumed. We can then examine the \gls{det} curves and pick the best threshold per subgroup for a certain \gls{fpr}.

We evaluated \gls{tar} for specific \gls{far} values. As described in Section~\ref{subsec:pf}, the verification experiments were 5-fold, with no overlap in subject ID between folds. Results reported are averaged across folds in all cases and are shown in Table~\ref{tab:ethnicy-far}. For each subgroup, the \gls{tar} of using a global threshold is reported (upper row), as well as using the optimal per subgroup threshold (lower row). 

Even for lower \gls{far}, there are notable improvements, often of the order of 1\%, which can be challenging to achieve when \gls{far} is near $\geq$90\%. More importantly, each subgroup has the desired \gls{fpr}, so that substantial differences in \gls{fpr} will remain unfounded. We experimented with ethnicity estimators on both the query and test image, which yielded similar results to those reported here.

\begin{table}[t!]
\begin{center}
    \caption{\small{\textbf{Human assessment (quantitative).} Subgroups listed per row (\ie human) and column (\ie image). Note, most do the best intra-subgroup (\textcolor{blue}{blue}), and second-best (\textcolor{red}{red}) intra-subgroup but inter-gender. WF performs the best; WF pairs are most correctly matched.}}
    \label{tab:humsn-eval-results} 
     \vspace{-2mm}
\footnotesize
\scalebox{0.94}{
\begin{tabular}{c}
\begin{tabular}{c l c c c cc}
&&\multicolumn{4}{c}{\textbf{Image}}\\
   \multicolumn{2}{c}{(Acc, \%)}& CF  & CM & WF &WM& Avg\\
\end{tabular}\\
\begin{tabular}{c l| r r r r| r}
\cline{3-7}
       &CF &\  \textbf{\textcolor{blue}{52.9}}&   \textbf{\textcolor{red}{48.0}}&43.8&44.7 &47.4 \\
       
        \parbox[t]{2mm}{\multirow{3}{*}{\rotatebox[origin=c]{90}{\textbf{Human}}}}  \hspace{-5mm}
        
        &CM &  \textbf{\textcolor{red}{45.6}} & \textbf{\textcolor{blue}{50.4}}  & 44.4 &36.2 &44.1 \\
       
        &WF & 44.7 &43.8 & \textbf{\textcolor{blue}{57.3}}& \textbf{\textcolor{red}{48.0}} & \textbf{48.5} \\
        &WM & 30.1&\textbf{\textcolor{red}{47.4}} &  45.3 & \textbf{\textcolor{blue}{56.1}} & 44.7\\\cline{3-7}
        &Avg &  43.3& 47.4&\textbf{47.7} &46.3 &46.2\\
 \end{tabular}
 \end{tabular}
 }
 \end{center}
\vspace{-5mm}
\end{table} 

\begin{figure}[t!] 
	\centering    
	\includegraphics[width=.55\linewidth] {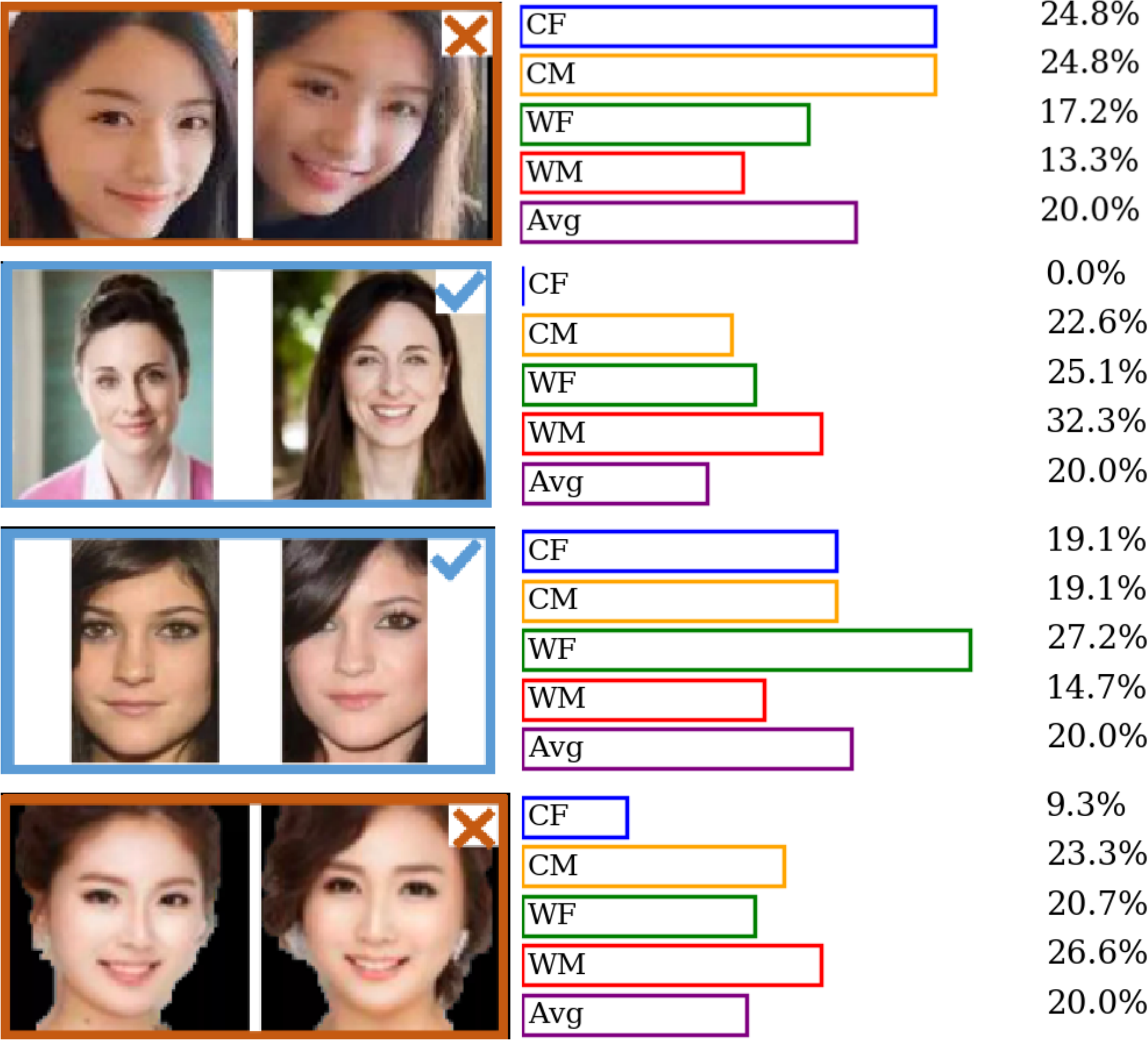}
		\caption{\small{\textbf{Human assessment (qualitative).} $\checkmark$ for \emph{match}; $\times$ for \emph{non-match}. Accuracy scores shown as bar plots. Humans are most successful at recognizing their own subgroup, with a few exceptions (\eg bottom).}}
		\label{fig:human-eval} 
		 \vspace{-5mm}
\end{figure}

\vspace{-5mm}
\noindent\paragraph{Human evaluation analysis.}
Subjects of a subgroup likely have mostly been exposed to others of the same (Table~\ref{tab:humsn-eval-results} and Fig.~\ref{fig:human-eval}). Therefore, it is expected they would be best at labeling their own, similar to the same ethnicity, but another gender. Our findings concur. Each subgroup is best at labeling their type, and then second best at labeling the same ethnicity but opposite sex. 

Interestingly, each group of images is best tagged by the corresponding subgroup, with the second-best having the same ethnicity and opposite gender. On average, subgroups are comparable at labeling images. Analogous to the \gls{fr} system, performance ratings differ when analyzing within and across subgroups. In other words, performance on \gls{bfw} improved with subgroup-specific thresholds. Similarly, humans tend to better recognize individuals by facial cues of the same or similar demographics. Put differently, as the recognition performances drop with a global threshold optimized for one subgroup and deployed on another, human performance tends to fall when across subgroups (\ie performances drop for less familiar subgroups).


\glsreset{bfw}
\section{Conclusion}
We introduce the \gls{bfw} dataset with eight subgroups balanced across gender and ethnicity. With this, and upon highlighting the challenges and shortcomings of grouping subjects as a single subset, we provide evidence that forming subgroups is meaningful, as the FR algorithm rarely makes mistakes across subgroups. We used an \textit{off-the-shelf} Sphereface, hypothesizing this SOTA CNN suffers from bias because of the imbalanced train-set. Once established that the results do suffer from problems of bias, we observed that the same threshold across ethnic and gender subgroups leads to differences in the \gls{fpr} up to a factor of two. Also, we clearly showed notable percent differences in ratings across subgroups. Furthermore, we ameliorate these differences with a per-subgroup threshold, leveling out \gls{fpr}, and achieving a higher \gls{tpr}. We hypothesized that most humans grew among more than their own demographic and, therefore, effectively learn from imbalanced datasets. In essence, a human evaluation validated that humans are biased, as most recognized their personal demographic best. This research, along with the data and resources, are extendable in vast ways. Thus, this is just a sliver of the larger problem of bias in ML.

{\small
\balance
\bibliographystyle{ieee_fullname}
\bibliography{references}
}


\end{document}